\documentclass[11pt,a4paper]{article}
\usepackage[hyperref]{emnlp2020}
\usepackage{times}
\usepackage{latexsym}

\usepackage{algorithm,amsmath}
\usepackage{pifont}
\usepackage{bbding}
\newcommand{\xmark}{\ding{55}}%

\usepackage[noend]{algorithmic}
\usepackage{url}
\usepackage{graphicx}
\usepackage{microtype}
\aclfinalcopy 
\usepackage{xcolor}
\usepackage{csquotes}
\usepackage{booktabs}

\title{Knowledge-guided Open Attribute Value Extraction with Reinforcement Learning}

\author{$\textbf{Ye Liu}^{1}$\thanks{\ \ Ye Liu and Sheng Zhang contributed equally.} \ \ $\textbf{Sheng Zhang}^{1}$\footnotemark[1]  \ \ $\textbf{Rui Song}^{1}$\thanks{\ \ Rui Song and Yanghua Xiao are corresponding authors. Yanghua Xiao was supported by Shanghai Science and Technology Innovation Action Plan (No.19511120400).}  \ \ $\textbf{Suo Feng}^2$\ \ $\textbf{Yanghua Xiao}^{2}$\footnotemark[2]\\
  $\textbf{ }^1$Department of Statistics,
  North Carolina State University,
  Raleigh, NC 27695-8206 \\
  $\textbf{ }^2$Shanghai Key Laboratory of Data Science, School of Computer Science, Fudan University, China 200433\\
  \texttt{\{yliu87, szhang37, rsong\}@ncsu.edu}, \ \ \texttt{\{fengs17, shawyh\}@fudan.edu.cn}  \\
  
  }

\date{}

\begin{document}
\maketitle
\begin{abstract}
Open attribute value extraction for emerging entities is an important but challenging task. 
A lot of previous works formulate the problem as a \textit{question-answering} (QA) task. 
While the collections of articles from web corpus provide updated information about the emerging entities, the retrieved texts can be noisy, irrelevant, thus leading to inaccurate answers.
Effectively filtering out noisy articles as well as bad answers is the key to improving extraction accuracy. Knowledge graph (KG), which contains rich, well organized information about entities, provides a good resource to address the challenge.
In this work, we propose a knowledge-guided reinforcement learning (RL) framework for open attribute value extraction. 
Informed by relevant knowledge in KG, we trained a deep Q-network 
to sequentially compare extracted answers to improve extraction accuracy.
The proposed framework is applicable to different information extraction system.
Our experimental results show that our method outperforms the baselines by 16.5 - 27.8\%. 

\end{abstract}
\section{Introduction}
Numerous entities are emerging everyday. The attributes of the entities are often noisy or incomplete, even missing. 
In the field of electronic commerce, target attributes (e.g., brand, flavor, smell) of new products are often missing \cite{zheng2018opentag}. 
In medical analysis, attributes like transmission, genetics and origins of a novel virus are often unknown to people. 
Even in DBpedia, a well-constructed and large-scale knowledge base extracted from Wikipedia, half of the entities contain less than 5 relationships \cite{shi2018open}. 
A method that is capable of supplementing reliable attribute values for emerging entities can be highly useful in many applications. 

\begin{figure*}[ht]
  \centering
  \includegraphics[width=1\textwidth]{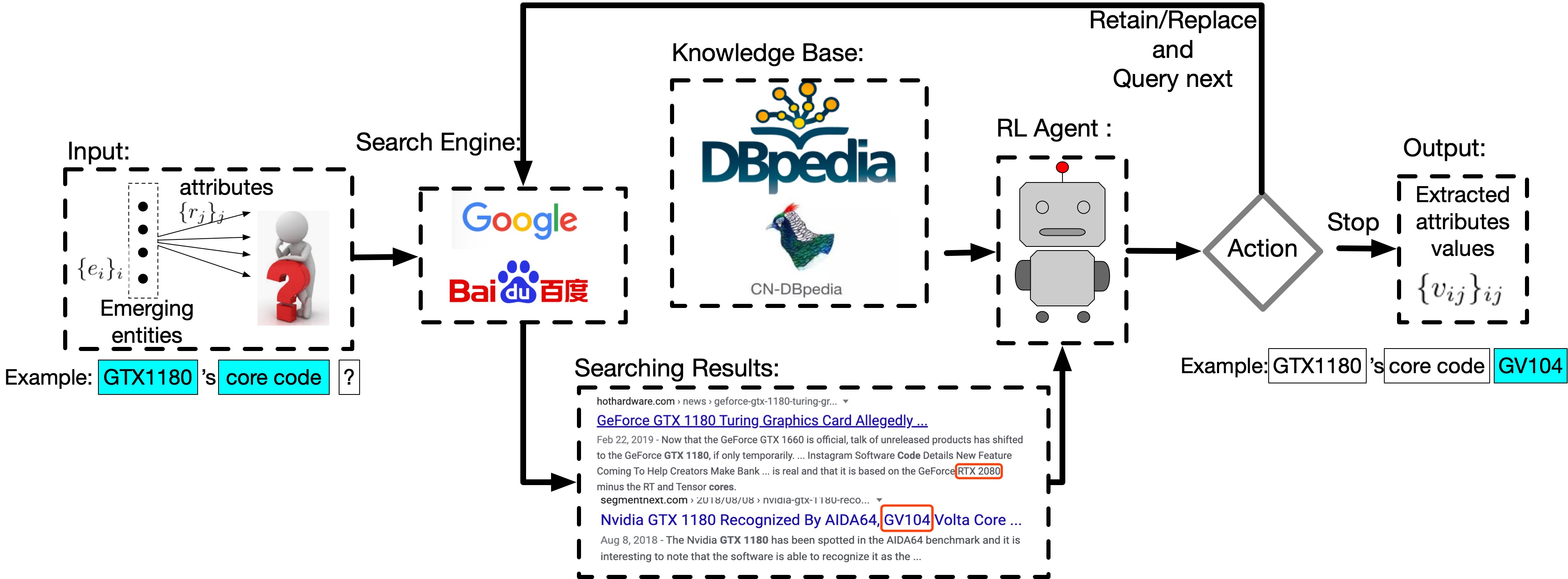}
 \caption{Illustration of overall process. The inputs are pairs of entities and attributes.
 Relevant articles are retrieved via search engines. The articles together with the KG are fed into the RL agent to inform the selection between candidate answers and the stopping decisions. When the RL agent decides to stop, it will output the best extracted answer.
}
 \label{fig:1}
\end{figure*}

Although information extraction methods have been extensively studied, the task of open attribute value extraction remains challenging.
First, the emerging entities may have new attribute values that are absent in the existing KG. Under such circumstances, the prediction methods under the closed-world assumption and the methods that cannot utilize external information are not well suited due to their limited recalls.
Second, 
while web corpus can be used as a good resource to provide relatively updated and relevant articles for large varieties of emerging entities,
the articles retrieved from web corpus can be noisy and/or irrelevant, which in turn leads to a limited precision. 
Finally, even when articles are relevant, the extracted answers might still be inaccurate due to the error-prone information extraction model.

To effectively filter out noisy answers that are obtained either due to the irreverent articles or the errors incurred by the information extraction system, we 
pose the following two questions: First, how many articles should we collect from the enormous web corpus?
Second, 
how to select the most reliable value out of the pool of all the possible answers extracted from the articles?

There is no common answer to the first question that works for all triplets
because of the inconsistent degrees of difficulties in finding the correct attribute values.
The decision of when to stop querying more external articles needs to be made after successive evaluations of the candidate answers.
Thus the decision making process is inherently sequential.

Reinforcement learning (RL) is a commonly adopted method to deal with sequential decision problems and has been widely studied in the field of robotic and game \cite{sutton1998introduction}. But there are not many researches on open attribute value extraction with RL. 
One existing literature of RL-based method for value extraction is proposed by \cite{narasimhan2016improving}. 
In their work, a RL framework is designed to improve accuracy of event-related value extraction by acquiring and incorporating external evidences. 
However, their approach requires a great amount of context information about the specific event of interest
during the training process. 
It is not trivial to extend their framework for open attribute value extraction, because we would need to collect context words and train a new model with annotated data for each emerging attribute.
Therefore, their framework cannot be generalized to open attribute value extraction task when
various entities and attributes are involved.

While using the context words to construct the states in RL is not suitable in our task, 
our solution is to leverage the rich, well-organized information in KG, which is not only informative but also generalizable. 
Such information can be leveraged in answer comparisons, which addresses our second question.
For example, to fill the incomplete triplet $<$ \textit{iPhone 11}, \textit{display resolution}, ?$>$, from the KG we may find that the attribute values  ``display resolutions" of an entity that is under category ``Phone" is commonly expressed in the format of ``xxx by xxxx Pixels", where x stands for some digit. The typical instances of the attribute values for entities under the same category provide valuable background information. 

In this paper, we propose a knowledge-guided RL framework to perform open attribute value extraction. 
The RL agent is trained to make good actions for answer selection and stopping time decision. 
Our experiments show that the proposed framework significantly boosts the extraction performance.

To the best of our knowledge, we are the first to integrate KG in a RL framework to perform open attribute value extraction
In summary, our contribution are in three folds: 
\begin{itemize}
\itemsep0em 
    \item We construct a novel knowledge-guided RL framework for open attribute value extraction task.
    \item We provide a benchmark data set for open attribute value extraction task. 
    \item Our method achieves a significantly better performance than the state-of-the-art methods.
\end{itemize}

\section{Overview}\label{sec:overview}
\textbf{Problem Definition}

We denote the entity-attribute-value triplet as $<e, r, v>$. The goal is to find the attribute value in an incomplete triplet $<e, r, ?>$.
To achieve this purpose, we pose a question generated with a pre-defined template to search engine to obtain relevant articles.
For example, to fill the incomplete triplet $<$ \textit{GTX1080}, \textit{Core code}, ?$>$, we retrieve articles with the query \enquote{What is the core code of GTX1080?}.

An information extraction system, such as QANet \cite{yu2018qanet}, is used to extract a candidate answer with a certain confidence score from an article.
However, due to the inconsistent qualities of the online articles and the inevitable errors caused by the information extraction system,
the results of extracted from only one online article is not satisfactory in many cases.
Another source of information that can be leveraged help to fulfill such a task is a KG. 
While it is hard to find out the attribute values for an emerging entity given the existing ones, 
the KG can serve as the background knowledge about the attributes. 
We approach the problem using a reinforcement learning framework that is illustrated in the next section.

\begin{figure*}[ht]
  \centering
  \includegraphics[width=1.05\textwidth]{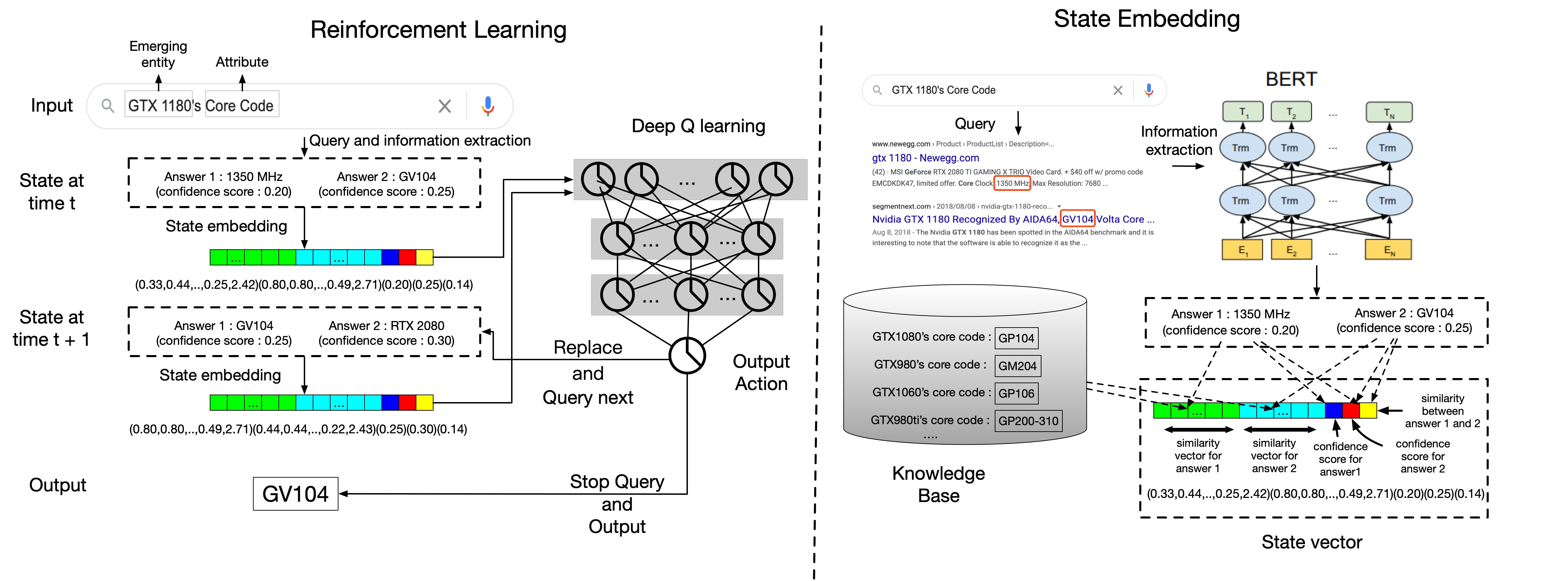}
     \caption{An example of reinforcement learning framework and state embedding. On the left panel, the inputs are the emerging entities and attributes. \textit{The current best answer} and \textit{the next candidate answer} are extracted from the retrieved articles with an information extraction system. The extracted answers are embedded into a state vector from the state embedding process. The state vector is fed into a policy network (DQN). The policy network selects the optimal actions and outputs the current best result. On the right panel, it presents the state embedding process. The extracted candidate answers are embedded into a state vector via similarity metrics and confidence scores.}
 \label{fig:2}
\end{figure*}

\textbf{System Overview}

Our procedure is summarized in Figure \ref{fig:1}.
We use $<$\textit{GTX1180}, \textit{core code}, ?$>$ as an example for illustration.
The query \enquote{What is the core code of GTX1180?} is posed
to the search engine to obtain a collection of relevant articles
by downloading the top $M$ headlines and bodies in the searching page.
$M$ is a pre-determined parameter that controls the maximum capacity of the retrieved articles.
For each of the retrieved articles, we use an information extraction system to extract a candidate answer.
In our example, \textit{RTX2080} is extracted with a confidence score of 0.30 from the first queried article and \textit{GV104} is extracted with a confidence score of 0.25 from the second article. 
Given the first two candidate answers, the RL decides on which answer to pick and whether more articles need to be retrieved.

To make such decisions, in addition to the confidence evidence from the information extraction system, the relevant facts in the KG will be fed into the RL agent to serve as the background knowledge about the attribute.  
For a triplet $<e, r, ?>$, we consider
$v^r$ as a reference value with respect to the attribute $r$ if there is a triplet
$<e', r, v^r>$ and $e$, $e'$ belong to the same category\footnote{The category information is obtained from concept of CN-DBpedia. The knowledge base contains multi-level hierarchy of categories. We use the lowest-level (most specific) category in the hierarchy to derive the reference values.} in the KG. 
In our example, since \textit{GTX1180} belongs to the category
\textit{NVIDIA GPU}, and so does 
\textit{GTX1080} and \textit{GTX980},
the reference values are retrieved from the fact that
the \textit{core code} of \textit{GTX1080} is \textit{GP104} and 
the \textit{core code} of \textit{GTX980} is  \textit{GM204}.
Guided by the KG, the RL agent makes successive evaluations and finally outputs the predicted candidate attribute value 
via a policy network such as DQN \cite{mnih2015human}.

\section{Reinforcement Learning for Open Attribute Value Extraction}

The attribute value extraction task is modeled as a \emph{Markov decision process} (MDP),
where the RL agent is actively engaged in the decision making process to maximize the reward, which measures the correctness of the extracted attribute values. 

 The MDP is modeled as a tuple $(S, A, T, R)$, where
 $S = \{s\}$ is the space of all possible real-valued vector states;
 $A=\{a\}$ is the set of actions;
 $T(s'|s,a)$ refers to a transition function that maps the domain of state and action to a probability distribution of states;
 $R(s,a)$ is a reward function that maps the domain of state and action to a real number,
 which is encoded such that the higher value the better.
 We describe our RL methodology by illustrating these components as follows.

\textbf{Action and transition}
At each decision stage, the agent will observe two candidate answers from two articles and make decisions to answer the two questions: (i) which answer is better out of the two? (ii) should the agent stop at \textit{the current best answer} or continue querying more articles? At the initial decision point, two candidate answers are obtained 
from two articles simultaneously queried from the web,
where we arbitrarily assign one of them to be the current best answer and the other the new candidate answer.


We define the following three actions in $A$:
\begin{enumerate}
    \item \textit{Retain}: (i) retain the current best answer and discard the new answer; (ii) query next.
    \item \textit{Replace}: (i) replace the current best answer with the new answer; (ii) query next.
    \item \textit{Stop}: (i) select the current best answer as the final answer; (ii) stop the query.
\end{enumerate}

At all subsequent decision points, we will retain or replace the \emph{current best answer} and continue comparing with the \emph{new candidate answers} queried from the web until the action is \textit{``Stop"}. 

\textbf{State} At each decision point, the state is constructed by concatenating the following three components, where different sources of information are combined.
 
\textbf{(1) State variables associated with the confidence scores.}
The first component is the confidence scores associated with the two candidate answers,
which are defined by the information extraction system.
We consider this part as the signal of the goodness of the extracted answers related to the articles.

\textbf{(2) State variables informed by the KG.}
The second component leverages knowledge from the reference values. 
For a given attribute, we expect the attribute values to be similar to each other in lexical sense. 
In order to capture such information, we first construct 7 features based on 2 string lexical similarity metrics. 
For each of the 7 features, we take the average and maximum of the features for each of the two candidate answers as state variables.

\paragraph{String similarity metrics} The two string lexical similarity metrics as follows:
        $$L\_Sim(s_1, s_2) = 1 - \frac{L(s_1, s_2)}{max(|s_1|,|s_2|)},$$
    $$LCS\_Sim(s_1, s_2) = \frac{|LCS(s_1, s_2)|}{max(|s_1|,|s_2|)},$$
where $L(s_1, s_2)$ refers to Levenshtein distance \cite{levenshtein1966binary}. It measures how different two strings are by counting the number of deletions, insertions or substitutions required to transform $s_1$ into $s_2$. $L\_Sim(s_1, s_2)$ is known as the Levenshtein similarity and \emph{LCS}($s_1$,$s_2$) stands for the longest common sub-string of $s_1$ and $s_2$ \cite{gusfield1997algorithms}. 


\paragraph{Features based on similarity} We define the following 7 features to capture the similarity between two strings from different aspects:
\begin{itemize}
\itemsep0em 
    \item  $f_1$: \emph{L\_Sim} between $s_1$ and $s_2$;
    \item  $f_2$: \emph{LCS\_Sim} between $s_1$ and $s_2$;
    \item $f_3$: \emph{L\_Sim} between $s_1$ and $s_2$ with numbers removed from $s_1$, $s_2$;
    \item $f_4$: \emph{LCS\_Sim} between $s_1$ and $s_2$ with numbers removed from $s_1$, $s_2$;
    \item $f_5$: \emph{L\_Sim}  between $s_1$ and $s_2$ with  $s_1$, $s_2$ wildcard masked\footnote{Wildcard mask means masking the numbers in the string. For example, a string ``750 by 1334 Pixels" will be masked to ``xxx by xxxx Pixels".};
    \item $f_6$: \emph{LCS\_Sim} between $s_1$ and $s_2$ with  $s_1$, $s_2$ wildcard masked;
    \item $f_7$: The difference in the length of $s_1$ and $s_2$ in characters.
\end{itemize}


\paragraph{Construction of state variables by the KG} For each of the 7 features, given the reference values in $V^r$ and the two candidate answers, \textit{answer}\textsubscript{1} and \textit{answer}\textsubscript{2}, 
we form the 28 state variables in this part by taking averages and maximums, which is specified as follows:
\[
\begin{cases}
&\frac{1}{|V^r|}\sum_{\ell=1}^{|V^r|} f_i(\textit{answer}\textsubscript{1}, v^r_{\ell}),\\ 
  &\max_{\ell=1}^{|V^r|}  f_i(\textit{answer}\textsubscript{1}, v^r_{\ell}),\\
  &\frac{1}{|V^r|}\sum_{\ell=1}^{|V^r|}
  f_i(\textit{answer}\textsubscript{2}, v^r_{\ell}),\\  
  &\max_{\ell=1}^{|V^r|}
  f_i(\textit{answer}\textsubscript{2}, v^r_{\ell}),
  \end{cases}
\]
for $i = 1,\ldots,7$. This is the part of the state where knowledge from KG is used to inform the decision of the RL agent.

\textbf{(3) State variables based on the candidate answers}
The third component contains the Levenshtein similarity between the two candidate answers. Intuitively, when the confidence scores of both candidate answers are high and they are very similar to each other, then it shows some positive signal for stopping.

The components (1) - (3) are concatenated together to construct the 31-dimensional state vector to carry information from different perspectives.

\textbf{Reward}
The reward is set to 0 when the query process is ongoing; 
only at the final stage when the query is terminated, a nonzero reward is received,
which measures the similarity between the final answer and the correct answer.
\begin{equation*}
R(s,a) = \begin{cases}
L\_Sim(\hat{v}, v) & a \text{ is \textit{Stop}},\\
0 & \text{otherwise},
\end{cases}
\end{equation*}
where $\hat{v}$ is the selected best attribute value and $v$ is the true attribute value. 

\textbf{Method}
Since the state defined in our framework is from a continuous space, we adopt a deep Q-network (DQN) to approximate $Q(s, a)$ with a deep neural network denoted by $Q(s,a;\theta)$. 
Specifically, we parameterize an approximate value function $Q(s,a;\theta)$ using a three-layer deep neural network. The network takes the continuous 31-dimensional state vector $s$ as input and predict $Q(s, a)$.
We use the rectified linear unit (ReLU) activation functions in the hidden layers. The architecture is illustrated in Figure \ref{fig:2}. 

Algorithm \ref{algotable} provides complete details of our MDP framework for the DQN training phase.

\begin{algorithm}[p]
      \caption{The full details of our training Phase for the DQN agent with $\epsilon$-greedy exploration.}
  \begin{algorithmic}[1]
  \STATE Initialize a set of training triplets 
  
   $x_i=<e_i,r_i,v_i>\in X$ 
   \STATE Initialize parameters $\theta$ randomly
   \STATE Initialize replay memory $\mathcal{D}$ 
 
     \FOR{$x_i \in X$}
     	\STATE Download M articles by searching with query ``$[e_i]'s [r_i]$"
     	\STATE Queue the downloaded articles in $C_{i}$ 
	    \STATE Identify reference values from the KG and save them in $V^r_i$
	 \ENDFOR 
	
    \FOR{epoch = 1, \ldots,E}
   	 \FOR{$i = 1,\ldots, |X|$}
	 	\STATE Pop the first two articles in $C_i$ and obtain \textit{answer}\textsubscript{1} with \textit{confidence}\textsubscript{1} and
	 	\textit{answer}\textsubscript{2} with \textit{confidence}\textsubscript{2}
		\STATE Form the state $s_1$ given  \textit{answer}\textsubscript{1}, \textit{confidence}\textsubscript{1},
	 	\textit{answer}\textsubscript{2}, \textit{confidence}\textsubscript{2},
	 	$V^r_i$
		 \FOR{$t = 1,\ldots, M-1$}
		    \STATE With probability $1-\epsilon$ select $a_t = \arg\!\max_a Q(s_t, a; \theta)$ 
            otherwise select $a_t$ randomly
		 \IF {$a_t$ is \NOT \textit{\enquote{Stop}}} 
		 \STATE $r_t \leftarrow 0$
		        \STATE Pop next article from $C_{i}$ and 
                 obtain \textit{answer}\textsubscript{new} with \textit{confidence}\textsubscript{new} 
		    \IF{$a_t$ is \textit{\enquote{Retain}}}
    			\STATE \textit{answer}\textsubscript{2}	 $\leftarrow$  \textit{answer}\textsubscript{new} 
    			\STATE \textit{confidence}\textsubscript{2} $\leftarrow$  \textit{confidence}\textsubscript{new}
            \ENDIF
            \IF{$a_t$ is \textit{\enquote{Replace}}}
                 	\STATE \textit{answer}\textsubscript{1}	 $\leftarrow$  \textit{answer}\textsubscript{2} 
        			\STATE \textit{confidence}\textsubscript{1} $\leftarrow$  \textit{confidence}\textsubscript{2}
        			\STATE \textit{answer}\textsubscript{2}	 $\leftarrow$  \textit{answer}\textsubscript{new} 
        			\STATE \textit{confidence}\textsubscript{2} $\leftarrow$  \textit{confidence}\textsubscript{new}
		      \ENDIF
    		\STATE Form a new state $s_{t+1}$
    	\ELSE 
    	        \STATE $\hat{v}_i \leftarrow \textit{answer}\textsubscript{1}$
        		\STATE $r_t$ = \textit{L\_Sim}($\hat{v}_i$, $v_i$)
        		\STATE $s_{t+1} \leftarrow NULL$
        \ENDIF
		\STATE Store transition $(s_t, a_t, r_t, s_{t+1})$ in $\mathcal{D}$ 
		\STATE Sample random mini batch of transitions $(s_t, a_t, r_t, s_{t+1})$ from $\mathcal{D}$ 
		\STATE $y_t = r_t$ \ \algorithmicif\ $a_t$ is \textit{"Stop"}
		
		\algorithmicelse\
		$r_t + \gamma max_{a'} Q(s_{t+1}, a'; \theta)$
		\STATE Update parameter $\theta$ on the loss $\mathcal{L}(\theta) = (y_{t} - Q(s_t, a_t; \theta))^2$
		\STATE   \algorithmicif\ {$a_t$ is \textit{``Stop"}}  \algorithmicthen\ break 	 
	    	\ENDFOR
     	 \ENDFOR
	 \ENDFOR
  \end{algorithmic}

      \label{algotable}
\end{algorithm}

\section{Experiments}

In this section, we compare our proposed RL framework to the state-of-the-art extraction-based baselines, demonstrating its robustness and ability to obtain accurate answers for missing attribute values. Our codes are publicly available online.\footnote{https://github.com/yeliu0930/Knowledge-guided-Open-Attribute-Value-Extraction-with-Reinforcement-Learning}

\subsection{Data}

The dataset is generated from existing triplets using the largest public Chinese knowledge base, CN-DBpedia \cite{xu2017cn}, with a corresponding taxonomy CN-Probase. \footnote{As far as we know, there is no public benchmark dataset suitable for our open attribute value extraction task with labeled values.}
Specially, the number of training triplets is 1022. The selected entities in the experiment are from four different fields, including GPU, game, movie and phone.
The testing data contains 75 triplets for each field, hence the total number of triplets in the testing is 300.
For each triplet in the training and testing data, we download articles from top $M=10$ links obtained from the Baidu search engine.
The CN-DBpedia is used as our external KG with the triplets in training and testing masked and to provide reference values.



\subsection{Reinforcement Learning Implementation}

In the RL setting, we use DQN to train the policy. Specifically, the DQN contains three layers of multilayer perceptron (MLP). The dimensions of hidden layers in MLP are chosen as 10 and 5 respectively. The dimension for the output is 3 which represents the three actions. 
In our experiments, the DQN model is trained 100 epochs where each epoch contains 1,000 transitions. 
We use a decreasing learning rate with epochs during the training process\footnote{The learning rate schedule is set to [[0, 0.05], [20, 0.01], [30, 0.005], [50, 0.001]].}.
The $\epsilon$ in $\epsilon$-greedy exploration is annealed from 1 to 0.02 over 10,000 transitions.
The replay memory $\mathcal{D}$ is of size 10,000.
We deploy our RL model in RLlib \cite{liang2017rllib} for efficiently distributed computation.



\subsection{Information Extraction System}
Different information extraction methods are implemented during the experiment. Sequence labeling methods including Bi-LSTM labeling, Bi-LSTM-CRF labeling \citep{huang2015bidirectional}, 
CNN labeling \citep{collobert2011natural}, CNN-att-CRF labeling \citep{tan2018deep}, and
OpenTag labeling \citep{zheng2018opentag} are used. 
We also consider three \textit{Machine Comprehension} (MC) models, including BiDAF \cite{seo2016bidirectional}, QANet
and a Bert-based model\cite{devlin2018bert}.
The SQuAD-like Chinese open-domain MC datasets including WebQA \cite{li2016dataset} and CIPS-SOGOU factoid question-answering subtask dataset \footnote{http://task.www.sogou.com/cips-sogou\_qa/} are used to train the extraction systems with distant supervision.

\subsection{Competitors}
We experimented with three traditional aggregation methods and four variants of RL agents as the competitors during the experiment.  

\textbf{Traditional Aggregation Methods}

1. Random choice (\textit{Random}): We randomly select an article out of $M$ articles and extract an answer from it with the information extraction system as the final answer.

2. First article (\textit{First}): We use the answer extracted from the article that ranked first in the search engines.

3. Majority aggregation (\textit{Majority}): We use a majority vote strategy over all the extracted answers.

4. Confidence aggregation (\textit{Confidence}): The answer with the highest confidence score out is chosen as the final answer. This aggregation method is only feasible when each candidate answer is associated with a confidence score.


\textbf{Variations of the RL framework}

1. \textit{RL-NK}: (No KG included) The RL agent do not leverage the information from KG. The KG-dependent part (i.e. the component (2) in state construction) is omitted from the state. 

2. \textit{RL-NR}: (No retain or replace actions)
The only action in the RL framework is \textit{Stop}. The final answer is the one with the highest confidence score among candidate answers seen before stop.


3. \textit{RL-NS}: (No stop action)
The RL agent do not make \textit{Stop} decisions. All of the M extracted candidate answers are compared.

4. \textit{RL-KG}: Our proposed RL framework.

Since the sequence labeling methods cannot provide valid confidence scores associated with the candidate answers, the answers extracted with these methods are aggregated using \textit{Random} and \textit{Majority} strategies.
For the MC models, we implemented all the aggregation strategies including our RL-based methods.

\subsection{Results}\label{sec:experiment_result}

Our evaluating metric is the Levenshtein similarity between the final answer
and the ground truth, which ranges from 0 to 1 and higher score represents better performance.
The results are summarized in Table \ref{table:3} when different
information extraction systems are combined with different aggregation strategies. 
The results are evaluated separately under each field and the combined results are also reported in the tables. All results reported are averaged over 3 independent runs. The oracle performances are provided to differentiate the error incurred by imperfect decisions and the inherent errors caused by the information extraction system. 

\begin{table}[H]
\resizebox{\columnwidth}{!}{%
\begin{tabular}{llllll}
           & \multicolumn{5}{c}{Evaluating Dataset} \\
           \cline{2-6}
          & GPU    & Games & Movie & Phone & All   \\ 
          \toprule
\multicolumn{6}{l}{\textbf{Baselines: Sequence labeling methods with traditional aggregations}} \\ [0.1cm]
\textit{Random}(OpenTag)  &0.222 &0.220& 0.238&0.244& 0.231 \\
\textit{Majority}(OpenTag) &0.282 &0.334 &0.321 & 0.320& 0.314 \\
\textit{Random}(Bi-LSTM)  &0.291 &0.322& 0.184&0.418& 0.304 \\
\textit{Majority}(Bi-LSTM)  &0.307 &0.334&0.194 &0.462& 0.324 \\
\textit{Random}(Bi-LSTM-CRF)  &0.349 &0.273& 0.287&0.336& 0.311\\
\textit{Majority}(Bi-LSTM-CRF)  &0.517&0.388& 0.360 &0.494& 0.440 \\
\textit{Random}(CNN)  &0.409 &0.333& 0.356&0.381& 0.370\\
\textit{Majority}(CNN)  &0.534&\textbf{0.419}& 0.430 &\textbf{0.508} & 0.473 \\
\textit{Random}(CNN-att-CRF)  &0.399 &0.272&0.304&0.310& 0.321\\
\textit{Majority}(CNN-att-CRF)  &\textbf{0.626}&0.413& \textbf{0.481} &0.506 &\textbf{0.507} \\[0.1cm] \midrule \midrule
\multicolumn{6}{l}{\textbf{BiDAF with traditional aggregations, RL methods and Oracle strategy. }} \\ [0.1cm]

\textit{Random}(BiDAF) &0.259  &0.155  &0.267  &0.223 & 0.226 \\
\textit{First}(BiDAF) &0.451  &0.498  &0.533  &0.632 & 0.528 \\
\textit{Majority}(BiDAF)  &0.488 & 0.321  & 0.539 & 0.415 &0.441 \\
\textit{Confidence}(BiDAF)  &\textbf{0.799}  & 0.488 & 0.645 & 0.560 &0.623\\ \hline
\textit{RL-NK}(BiDAF)  &0.679&0.602&0.609&0.658&0.637\\
\textit{RL-NR}(BiDAF)  &0.751&0.655&0.622&0.644&0.668 \\ 
\textit{RL-NS}(BiDAF)  &0.759 &\textbf{0.732} &0.673 &0.680 &0.711 \\
\textit{RL-KG}(BiDAF)  &0.786 & 0.692 &\textbf{0.686}  &\textbf{0.739} &  \textbf{0.726}\\
 \hline
\textit{Oracle}(BiDAF)  &0.902  &0.793  &0.846  &0.812  &  0.838 \\[0.1cm] \midrule \midrule
\multicolumn{6}{l}{\textbf{QANet with traditional aggregations, RL methods and Oracle strategy }} \\ [0.1cm]
\textit{Random}(QANet) & 0.261  & 0.167 & 0.259 & 0.236 & 0.230 \\
\textit{First}(QANet) & 0.507  & 0.533 & 0.531 & 0.675 & 0.561 \\
\textit{Majority}(QANet)      & 0.484  & 0.325 & 0.500 & 0.469 & 0.444 \\ 
\textit{Confidence}(QANet)   & 0.691  & 0.493 & 0.689 & 0.546 & 0.605 \\\hline
\textit{RL-NK}(QANet)    & 0.640  & 0.592 & 0.596 & 0.645 & 0.618\\ 
\textit{RL-NR}(QANet)  & 0.687  & 0.717 & 0.549 & 0.631 & 0.646  \\ 
\textit{RL-NS}(QANet)   & \textbf{0.801}  & \textbf{0.695} & 0.686 &0.771   &   0.738 \\ 
\textit{RL-KG}(QANet)  & 0.786  & 0.687 & \textbf{0.731} & \textbf{0.790} & \textbf{0.749} \\
 \hline
\textit{Oracle}(QANet)   & 0.932  & 0.840 & 0.878 & 0.868 & 0.880 \\ [0.1cm] \midrule \midrule 
\multicolumn{6}{l}{\textbf{BERT with traditional aggregations, RL methods and Oracle strategy}} \\ [0.1cm]
\textit{Random}(BERT) & 0.374  & 0.234 & 0.361 & 0.287 & 0.314 \\
\textit{First}(BERT) & 0.507  & 0.533 &0.531 &0.675 & 0.561\\
\textit{Majority}(BERT)   & 0.620  & 0.438 & 0.626 & 0.530 & 0.553 \\
\textit{Confidence}(BERT)     & 0.727  & 0.600 & 0.552 & 0.540 & 0.605 \\ \hline
\textit{RL-NK}(BERT)     & 0.716  & 0.682 & 0.565 & 0.687& 0.662\\ 
\textit{RL-NR}(BERT)   & 0.775 & 0.652 & 0.707 & 0.723 & 0.714  \\ 
\textit{RL-NS}(BERT)     & 0.773  & \textbf{0.673} & 0.769 & 0.831 & 0.762 \\ 
\textit{RL-KG}(BERT)     & \textbf{0.817}  & 0.637 & \textbf{0.777} & \textbf{0.837} & \textbf{0.767} \\
\hline
\textit{Oracle}(BERT)         & 0.925 & 0.857 & 0.887 & 0.909 & 0.895 \\\bottomrule
\end{tabular}
}
\caption{Accuracy of the baseline methods and our proposed methods. Bold indicates best baseline performances with a sequence labeling methods and best results achieved with BiDAF/QANet/BERT.
The Oracle performance shows the best possible performance when perfect decisions are made.
\textbf{Our proposed RL-KG improves the extraction performances substantially.} }
\label{table:3}
\end{table}

From Table \ref{table:3}, we have the following observations.
First, the RL based methods outperform all the competing baseline methods. 
By adopting the RL framework instead of traditional aggregation methods, the accuracies are boosted substantially.  
It demonstrate the effectiveness of the RL framework.
Second, compared to the RL framework without the guide of KG (\textit{RL-NK}), our proposed \textit{RL-KG} framework achieves significantly better results. This suggests that the KG does provide valuable information in the task of the attribute value extraction.
Third, the \textit{RL-KG} framework outperforms all the other variants of the RL framework. It shows that considering answer selection and stopping decisions at the same time achieves the best performances.

\begin{figure}[ht]
  \centering
  \includegraphics[width=0.5\textwidth]{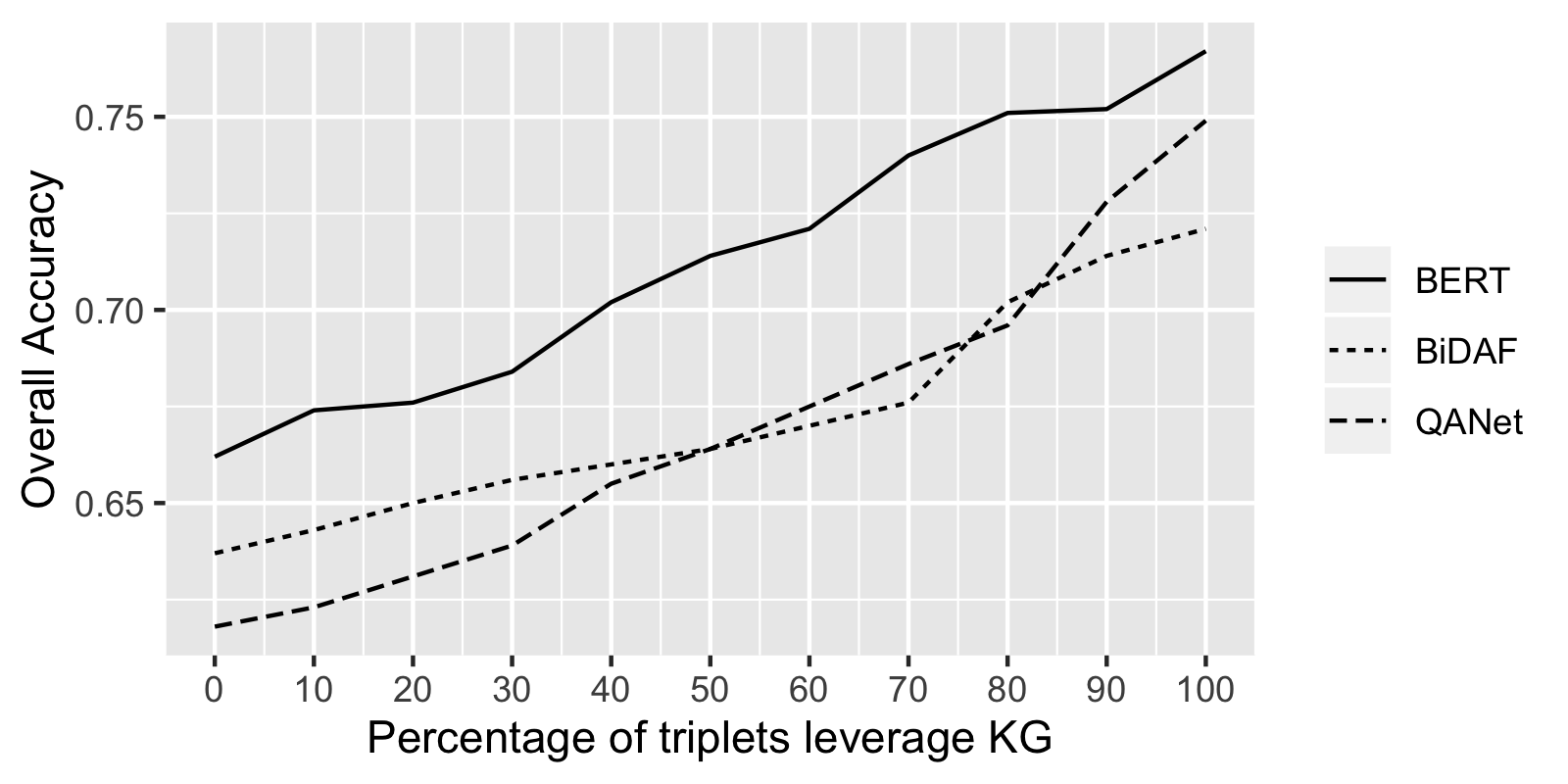}
  \caption{Extraction accuracy when KG is used with different levels of frequency. \textbf{Better extraction results are achieved when KG is used with higher frequencies.}}
  \label{fig:3}
\end{figure}

We also conduct an experiment to see how our method performs when the KG is not able to provide information for some triplets, which is a common situation in reality.  
During the experiment, we randomly set the reference values for $0-100$ (incremented by $10$) percent of triplets in the training and testing as empty. 
For those triplets that do not have reference values, the state variables associated with the KG (i.e. the component (2) in our state construction) are set to 0. 
Figure \ref{fig:3} displays how the performances change when information from KG is leveraged with different levels of frequency. 
It can be seen that for all the three information extraction models, the performances are getting better as the KG is used at higher frequencies.


\subsection{Case Study}

By incorporating information from KG, RL agent is able to rule out some unreasonable answers and boost the extraction accuracy.

\begin{table*}[ht]

\begin{tabular}{lllll}
Entity  & Attribute & Truth & \textit{RL-KG(BERT)}& \textit{Confidence(BERT)}\\  \hline
Founder r680-470   & 
Operating system & DOS & DOS & T2410\\
Super puzzles game  & Release date  & 12/25/2007  & 12/25/2007 & 43.75  \\ 
Madea's Witness Protection  & Rating system & USA:PG-13 & pg-13 & 6.4 \\
Tequila Works, S.L  & 
Number of endings & 2 & 2& ConsoleGame
\end{tabular}
\caption{Case studies (translated)}
\label{table:4}
\end{table*}

\begin{table*}
\begin{tabular}{l|l}
\hline \hline
Example 1 & \\ \hline
$<e,r,?>$  & $<$Founder r680-470,  Operating system, ? $>$ \\
...  & ... \xmark \\
Raw corpus 4   &  Intel Pentium dual-core {\color{blue}{T2410}} is an entry level processor based on the Merom-2M... \xmark\\
Raw corpus 5   &  ...Founder {\color{blue}{Q680}} high shot instrument scanner 5 million pixels A4 on... \xmark \\
Raw corpus 6  &  ...The latest R680 series laptop, Operating system {\color{blue}{DOS}} with ... \Checkmark \\ 
...  & ...\\
Reference values & $<$..IOS, DOS, Andriod, EMUI..$>$ \\ \hline \hline

Example 2 & \\ \hline
$<e,r, ?>$  & $<$Super puzzles game, Release date, ? $>$ \\
Raw corpus 1   & ...announced release date of super puzzles is {\color{blue}{12/25/2007}}. It was published by... \Checkmark\\
Raw corpus 2   &  ...Super Puzzle HD v2.1.2 -  Size: {\color{blue}{43.75}} MB...\\
...  & ...\\
Reference values & $<$..02/24, 03/27/2018, 05/07/2016..$>$ \\ \hline \hline
\end{tabular}
\caption{
Detailed examples where the trained RL agent helps to select the candidate answers from raw corpus. Blue words represent the extracted answers from BERT. The check-marks denote the selected answers.}
\label{table:5}
\end{table*}

In Table \ref{table:4}, we present some cases where the trained RL agent helps to correct the information extraction errors. 
More details for the first two examples are included in Table \ref{table:5}.
For the first example in Table \ref{table:5}, the emerging entity is \textit{Founder r680-470} and the attribute of interest is \textit{operating system}. The reference values retrieved from the KG include values like \textit{IOS, DOS, Andriod, EMUI}. The trained RL agent stops after querying the six raw corpus and selects the answer \textit{DOS} over the previous candidate values, which is exactly one of the reference values. 
In the second example in Table \ref{table:5}, we are interested in 
the \textit{release date} for 
the emerging entity \textit{super puzzles game}. The reference values are dates like \textit{02/24, 03/27/2018, 05/07/2016}, etc. The trained RL agent stops after one step, and outputs the candidate answer \textit{12/25/2007}.
By leveraging the reference values, our proposed method demonstrated its advantages in answer selection.

\section{Related Work}

\subsection{Machine reading comprehension}
Machine reading comprehension (MRC) and automated question (QA) answering
are important and longstanding topic in NLP research due to its huge potentials in wide variety of applications.
An end-to-end MRC QA models are expected to have the ability to read a piece of text and then answer questions about it. 
Significant progress has been made with the machine reading and QA task in recent years. Some notable works include BiDAF \citep{seo2016bidirectional}, 
SAN \citep{liu2017stochastic}, QANet,
ALBERT \citep{lan2019albert}.

Our proposed framework can also be regarded an end-to-end MRC QA model that is built on top of an existing MRC QA model, which is used as the information extraction system in our extraction process. 
Different from most of the previous works, our focus is to enhance the performance of an existing model by utilizing  external information from KG and by acquiring more articles when the agent does not feel confident about the extracted answer.

\subsection{Open-world knowledge graph completion}

Attribute value extraction under the open world assumption has received many attentions in NLP community recently.
There has been quite a few works on open attribute value extraction. 
OpenTag \citep{zheng2018opentag} formalized the extraction problem as a sequence tagging task and proposed an end-to-end framework for open attribute value extraction.
The open-world KGC \cite{shi2018open} used a complex relationship dependent content masking architecture to mitigate the presence of noisy text descriptions and extract the attribute value from the denoised text. 
TXtract \citep{karamanolakis2020txtract} incorporated the categorical structure into the value tagging system. 
However these methods suffer from irrelevant articles and is not able to filter out noisy answers. 

\subsection{NLP with reinforcement learning}
RL \cite{sutton1998introduction} is a framework that enables agents to reason about sequential decision making as an optimization process.
It has been widely applied in NLP tasks, including
article summarization \citep[][]{paulus2017deep,li2018actor,celikyilmaz2018deep},
dialogue generation \citep[][]{li2016deep, serban2017deep,li2019dialogue},
and question answering \citep{xiong2017dcn+,wang2018r,das2019multi} and so on.  
To the best of our knowledge, we are the first to integrate information from KG into a RL framework to fulfill the attribute extraction task.  

\section{Conclusion and discussion}

This paper presents a novel RL framework to perform open attribute value extraction. 
Through a set of experiments, we observe that the most of the computation cost is incurred by training the information extraction system. 
The remaining computation cost from RL framework is
comparably small during both the training and the prediction process. Specifically, during our experiments, we trained a three-layer deep neural network model, which has much fewer parameters compared to the information extraction system. 
The proposed RL method demonstrates promising performance, where the KG showed its ability to provide guidance in open attribute extraction task.
Our framework also contributes to areas of knowledge graph completion and automatic question-answering for attribute values. 

KG has huge potential to provide rich background information in many NLP applications. 
Our solution for attribute value extraction can be extended to other NLP tasks. 
A potential attempt might be to use KG to design the reward in the RL framework to provide weak supervision. 
We leave this as our future work.


\bibliography{KG_RL}
\bibliographystyle{acl_natbib}



\end{document}